\documentclass[a4paper, 10pt, conference]{ieeeconf}      

\IEEEoverridecommandlockouts                              

\usepackage{graphics}
\usepackage{amsmath}
\usepackage{amssymb}
\usepackage{balance}
\usepackage{mathtools}
\usepackage{hyperref}
\usepackage{xcolor}
\usepackage[font=footnotesize,hypcap=false]{caption}

\title{\LARGE \bf
mc-mujoco: Simulating Articulated Robots with \\ FSM Controllers in MuJoCo
}

\author{Rohan P. Singh$^{1,2}$, Pierre Gergondet$^{1}$, Fumio Kanehiro$^{1,2}$
\\\tt\small{Email: rohan-singh@aist.go.jp}
\thanks{
$^{1}$
CNRS-AIST JRL (Joint Robotics Laboratory) IRL,
National Institute of Advanced Industrial Science and Technology (AIST),
Japan.}
\thanks{
$^{2}$
University of Tsukuba,
Ibaraki,
Japan.}
}%

\begin{document}

\makeatletter
\let\@oldmaketitle\@maketitle%
\renewcommand{\@maketitle}{\@oldmaketitle%
    \centering
    \includegraphics[width=\textwidth,trim={0 0 0 0.15cm},clip]{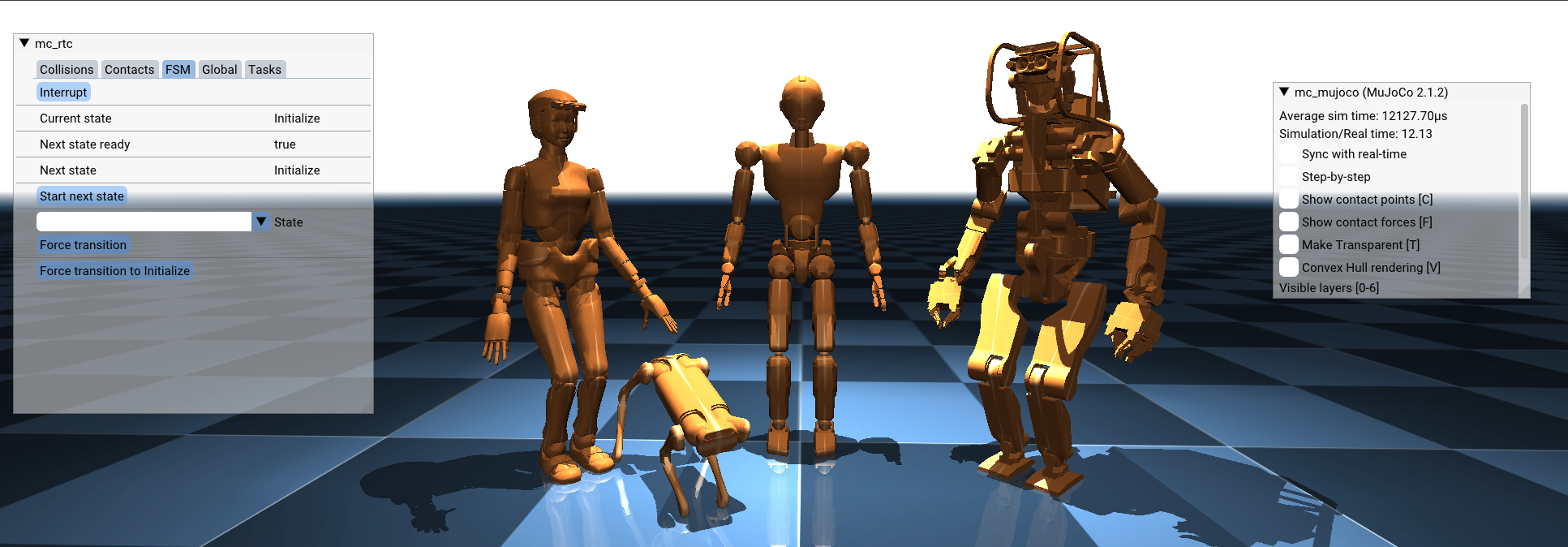}
    \captionof{figure}{A screengrab of the mc-mujoco interface with HRP-4CR (left), JVRC1 (middle),
    HRP-5P (right) humanoid robots, and the Unitree AlienGo quadruped being simulated simultaneously
    under a QP controller. The native \textit{mc-rtc} GUI and a new convenience panel are integrated
    into the same GLFW window eliminating the dependence on external applications. \textit{mc-rtc}
    visuals and 3D markers have been deliberately disabled.}
    \vspace{-.1cm}
    \label{figure:multi-robot-header}
}
\makeatother

\maketitle
\thispagestyle{empty}
\pagestyle{empty}

\begin{abstract}
For safe and reliable deployment of any robot controller on the real hardware platform,
it is generally a necessary practice to comprehensively assess the performance of the 
controller with the specific robot in a realistic simulation environment beforehand.
While there exist several software solutions that can provide the core physics engine
for this purpose, it is often a cumbersome and error-prone effort to interface the
simulation environment with the robot controller being evaluated. The controller
may have a complex structure consisting of multiple states and transitions within
a finite-state machine (FSM), and may even require input through a GUI. 

In this work, we present \textit{mc-mujoco} --- an open-source software framework that
forms an interface between the MuJoCo physics simulator and the \textit{mc-rtc} robot
control framework. We provide implementation details and describe the process for adding
support for essentially any new robot. We also demonstrate and publish a sample FSM controller
for bipedal locomotion and stable grasping of a rigid object by the HRP-5P humanoid robot in
MuJoCo. The code and usage instructions for \textit{mc-mujoco}, the developed robot modules,
and the FSM controller are available online.
\end{abstract}

\section{INTRODUCTION}
\label{section:intro}

With the open-sourcing of its source code, DeepMind's MuJoCo physics engine has become
a popular choice of simulation environment in the robotics research community. The library
claims to provide computationally efficient simulation without compromising significantly on
the accuracy and stability of the simulation \cite{todorov2012mujoco, mujocoblog, erez2015simulation}.
Specifically, it offers optimization-based contact dynamics that allow for soft contacts
and other constraints, in addition to using a generalized coordinates system. The library
offers a C API for interacting with the simulation environment, and the user must rely on
this API to read sensor measurements from the simulation state and to apply control signals
to the robot actuators, or to control other aspects of the scene like changing model parameters
or visualizations.

A powerful yet user-friendly software framework for implementing model-based controllers is
provided in \textit{mc-rtc} - a real-time middleware built on top of the SpaceVecAlg, RBDyn and
Tasks libraries \footnote{\url{https://jrl-umi3218.github.io/mc_rtc/index.html}}. \textit{mc-rtc}
has been used extensively to control several humanoids and robot arms using Quadratic Programming
(QP) \cite{escande2014hierarchical} based controllers to perform a diverse variety of demonstrations
\cite{cisneros2019qp}. It provides a friendlier way to implement complex robot behaviours along with
graphical tools for visualizations, user-inputs, logging, and more.

While the C API of MuJoCo provides a deep, low-level access to the simulator's functionality,
there is a need to develop a higher-level interface between \textit{mc-rtc} and MuJoCo so that
developers can reliably and efficiently simulate complex controllers and evaluate the robot's behavior
in a dynamic simulator before deployment on the real hardware. Ideally, such an interface would also
allow online user-interaction with the simulation environment to apply externally perturbation to the
robot or objects. We develop \textit{mc-mujoco} to satisfy exactly this requirement (\autoref{figure:multi-robot-header}).
\textit{mc-mujoco} supports any robot and controller expressed as \textit{mc-rtc} modules, and
eliminates the need for the user to directly interact with the low-level MuJoCo library. \textit{mc-rtc} interface is developed in a
transparent manner so that the same controller code used for simulation --- running in \textit{mc-mujoco} --- can also be used with the real
robot -- running in the appropriate interface for a given robot. This greatly simplifies the process of reliably assessing a controller.

Existing interface packages such as \textit{mc-openrtm} allow \textit{mc-rtc} to communicate with the
OpenRTM middleware \cite{ando2005rt} used in the HRP robots and Choreonoid simulator \cite{nakaoka2012choreonoid}.
\textit{mc-naoqi} is an interface developed for the Softback Robotics NAO and Pepper robots \cite{bolotnikova2021task}.
We argue that \textit{mc-mujoco} provides a higher level of versatility and unique tools compared to
the existing solutions along with simulation accuracy and speed due to the underlying MuJoCo physics.
Compared to using \textit{mc-openrtm}, our framework provides more flexibility due to independence
from OpenRTM. Moreover, MuJoCo has the ability to simulate joint damping, static friction (both unavailable in Choreonoid)
and rotor inertia --- important factors for achieving stable and realistic simulations.

To reiterate, we make the following contributions through this work:
\begin{itemize}
    \item We publish an open-source software framework \textit{mc-mujoco} \footnote{\url{https://github.com/rohanpsingh/mc_mujoco}}
    that interfaces a popular robot control framework \textit{mc-rtc} with the MuJoCo physics simulator, and consequently,
    allows simulation of a wide range of robots with FSM controllers.
    \item We publish robot description packages for:
    JVRC1 humanoid robot \footnote{\url{https://github.com/isri-aist/jvrc_mj_description}},
    the AlienGo quadrupedal robot \footnote{\url{https://github.com/rohanpsingh/aliengo_mj_description}},
    and achieve stable simulation in \textit{mc-mujoco}. We also perform simulations of bipedal locomotion with the
    HRP-5P and HRP-4CR humanoid robots.
    \item We develop an FSM controller for the HRP-5P humanoid robot and demonstrate its application for simulation of
    object grasping and manipulation using our proposed software framework \footnote{\url{https://github.com/rohanpsingh/grasp-fsm-sample-controller}}.
\end{itemize}


\section{APPROACH}

\begin{figure}[t]
  \centering
  \renewcommand{\thefigure}{2}
  \includegraphics[width=\linewidth]{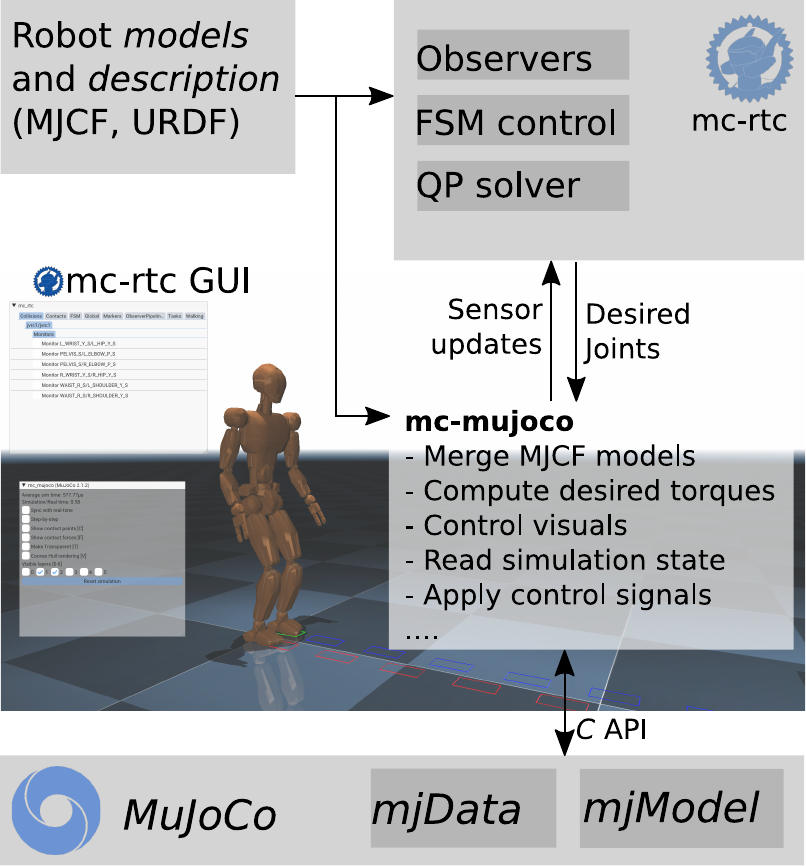}
  \caption{
  \footnotesize \textbf{Basic outline of the mc-mujoco software framework.}
  mc-mujoco acts as an interface between the low-level MuJoCo API and the
  \textit{mc-rtc} FSM controller. The native \textit{mc-rtc} GUI panel
  is embedded within the main window.
  }
  \label{figure:mc-mujoco-framework}
\end{figure}

In this section, we first describe the overall control framework and the key implementation details of
\textit{mc-mujoco}, and then outline the procedure to add support for new robots using description 
packages. Subsequently, we discuss features for multi-robot simulation and online user interaction using
the GUI. An overview of the proposed software architecture is shown in \autoref{figure:mc-mujoco-framework}.

\subsection{Interface overview}

\textbf {Actuator modelling and applying control signals.}
As it is common to find servo-type actuators driven by a PD control loop in most robots,
in \textit{mc-mujoco} we simulate the same effect by defining a direct-drive torque actuator
for each active joint in the XML model. The desired torque at the motor level is computed 
by a PD control loop and is then sent to these direct-drive actuators where it is scaled up
by specified transmission gear ratio by the MuJoCo engine and applied to the joint.

The input to the PD control loop --- reference joint positions and velocities --- are obtained
from the output of \textit{mc-rtc} computed by double-integration of the desired accelerations
predicted by the underlying QP controller.

The execution frequency of the PD controller is decided by the simulation timestep declared
in the model XML, usually set to 1 ms. Hence, the PD controller generally runs at 1000 Hz.
However, the mc-rtc controller could be running at a lower frequency, and consequently the
references to the PD control loop are interpolated from this lower frequency command signal. For model-based humanoid
locomotion, the closed-loop balancing controller is generally running at 200 Hz or more.

We also provide support for MuJoCo's \textit{position} and \textit{velocity} actuators.
Thus, a PD controlled actuator could also be created by attaching one \textit{position}
and one \textit{velocity} actuator to the same joint. Lastly, passive joints where
no actuator is attached are also supported in \textit{mc-mujoco}.

\textbf {Reading sensor data from the simulation state.}
In the opposite direction, feedback data from the simulation environment at each timestep
is read and relayed to \textit{mc-rtc}. This data includes measurements from sensors --- such
as Inertial Measurement Units (IMUs), force sensors, joint encoders, joint torque sensors ---
and ground truth values for position, orientation, linear and angular velocities, and linear
and angular accelerations of the floating-base link.

This data is conveyed to \textit{mc-rtc} which uses it to update state estimators and
provide feedback to the controller. The ground truth state of the floating-base link
from MuJoCo may be used for debugging purposes and to evaluate accuracy of the state
observer pipeline within the simulation environment, as this reading is rarely available
on a real robot without using complicated motion capture setups.

Further, we read pixels from all cameras defined in the robot model and render them in a
new window to simulate RGB cameras installed on the real robot. The RGB images
are also made available to the ``DataStore" facility of \textit{mc-rtc}, explained
below, and could be published over ROS in a Virtual Reality (VR) or teleoperation scenario.

\textbf {Datastore Calls.}
A powerful feature is provided by \textit{mc-rtc} to share arbitrary C++ objects --- including function objects --- between states
of an FSM or between the interface and the controller in the form of the ``DataStore" facility.
In \textit{mc-mujoco}, we use this feature to store callbacks for two purposes: (1) read
and write PD gains for the servos and (2) reading the 3-channel image matrix from a specified camera.
The callbacks may be called from within the controller code or any \textit{mc-rtc} client with
access to the controller's instance. The datastores for PD gains may be necessary for controllers
that use variable gain PD control --- where the gains are modified depending on the required robot
compliance or stiffness for a specific task. We plan to add support for depth images in a future
release.

\subsection{Robot description}
In the current implementation, any robot that is to be simulated in \textit{mc-mujoco} must be represented
in two structures - (1) the MJCF format for representing the robot in MuJoCo and (2) the \textit{robot module} for
representation in \textit{mc-rtc}. Environmental objects such as the ground plane or 3D objects for manipulation
do not need a \textit{robot module} representation if they do not need to be considered within the QP loop,
say, for obstacle avoidance.

\textbf{In MuJoCo.} The native MJCF format for modelling robots in MuJoCo can be easily derived from the more
popular URDF format. MJCF provides a wider range of features and model elements specific to MuJoCo, such as
options to select algorithms for contact computations and set parameters that affect the physics simulation.
It also allows for loading ``assets" such as meshes, height maps, textures and materials, that can later be 
referenced by other model elements. The MJCF files are generally stored on the disk as XML files (along with
meshes for collision and visualization), and are then parsed and compiled into a low-level data structure
called \textit{mjModel} using an internal compiler by MuJoCo.

\textbf{In mc-rtc.} In addition to the above, a robot must be represented as a \textit{RobotModule}
data structure within mc-rtc. The \textit{RobotModule} structure allows us to define additional information
about the robot such as the default joint configuration, collision pairs, names of sensors and devices attached
to the robot. Bounds for joint angles, joint velocities and torques can also be defined which could be used
by the controller to generate robot motion. Note that such bounds are also defined in the XML model file that
are enforced strictly internally by MuJoCo physics. But it is often useful to define additional, more conservative
bounds in the \textit{RobotModule} that could then be added as constraints to the QP in \textit{mc-rtc}.

With this work, we publicly release the robot description packages for the JVRC1 humanoid robot and the 
AlienGo quadrupedal robot. While creating the \textit{RobotModule} class for any robot is a straightforward
process, we plan to introduce a feature to automatically generate the robot module from the MuJoCo XML
model files. This will further simplify the process of adding support of new robots in \textit{mc-mujoco}.

\begin{figure*}
  \renewcommand{\thefigure}{3}
  \includegraphics[width=\linewidth]{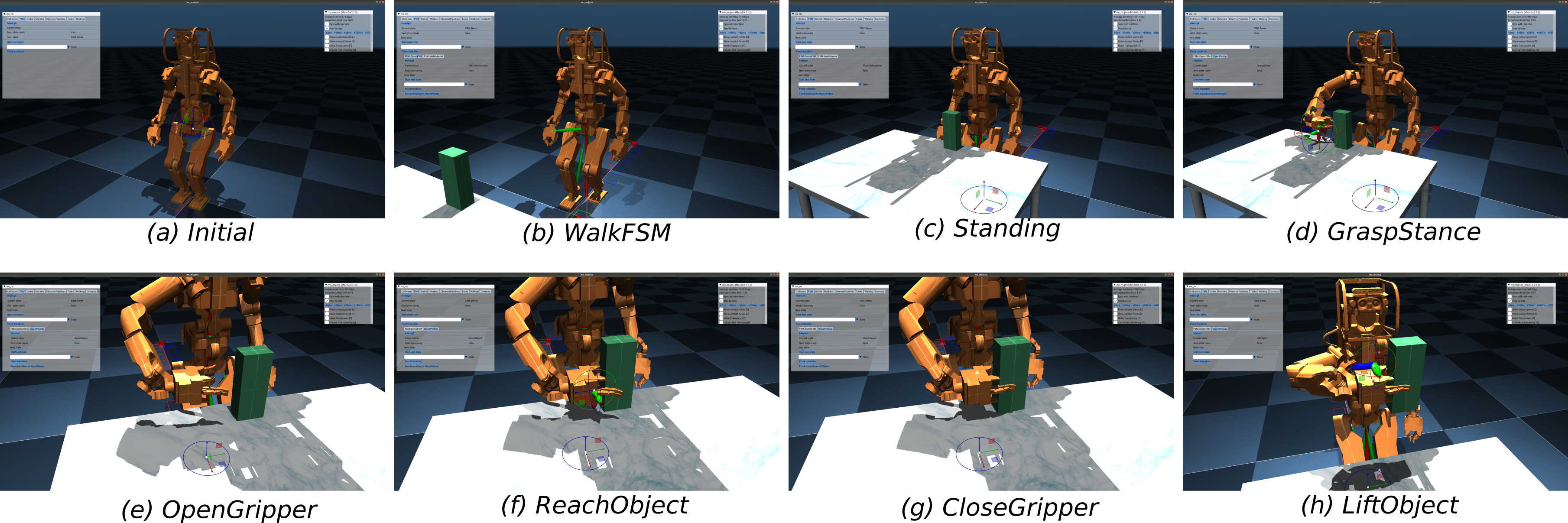}
  \caption{\textbf{FSM States and transitions.} Sample \textit{mc-rtc}
  controller for locomotion and object grasping.}
  \label{figure:fsm}
\end{figure*}

\subsection{User Interaction.}
We enable user interaction with the simulation in two crucial ways. Firstly, we allow the user to
apply external forces and external torques to any link of the robot (or, in fact, to any object)
with the use of mouse interactions. The ability to apply external perturbations is generally useful,
for example, to test robustness of control policies for humanoid locomotion or to displace 3D objects
in the environment to test the robustness of a reactive grasping algorithm with manipulator arms.
Secondly, we develop our own overlay Graphical User Interface (GUI) which allows the user to control
several elements of the simulation and visualization, like controlling the simulation speed, rendering
of contact points and contact forces, and toggle the visibility of ``geomgroups" --- generally used to separate visual elements from simulation elements --- of the robot model.

More importantly, we integrate the native GUI of \textit{mc-rtc} within \textit{mc-mujoco}. This is a
highly convenient feature because it allows the user to interact with the \textit{mc-rtc} controller
and the FSM framework to transition from one state to another, without launching another application
or adding dependency on ROS.
Using the built-in GUI the user can also tune controller parameters, such as PID tuning of joint
servos and visualize live plots from the controller. Further, interactive markers and visual handles
published by \textit{mc-rtc} are rendered in the scene which can be manipulated by the user, say, to
set the target position of a robot end-effector controller.

We use the popular DearImGui graphical user interface library for C++ \cite{imgui}
to develop both the GUIs within the \textit{mc-mujoco} window.

\subsection{Multi-robot simulation.}
Robot-to-robot interactions is a popular area of research and study including topics such
as multi-robot planning and cooperation, robot-robot object handover, and more. \textit{mc-rtc}
by itself provides support for multi-robot control by enabling constraints in the QP optimization problem
that account for all the loaded robots (for example, by adding appropriate collision
constraints, the QP can generate motion that avoids collision between 2 robots). In \textit{mc-mujoco},
multiple robots, each with their own controller, can be simulated simply by creating the description
files for each robot and then loading them through the \textit{mc-rtc} controller. \textit{mc-mujoco}
will automatically merge all the robot models into one MJCF model for representation in the MuJoCo
scene, while ensuring there is no name clashing. Objects in the environment can also be
loaded in the same manner by creating separate MJCF models for each object.

\autoref{figure:multi-robot-header} shows four robots being simulated simultaneously while being
controlled by a single \textit{mc-rtc} controller to hold the desired posture. The reference
position for any actuated joint of each of the robots can be changed through the GUI panel
which will lead the QP to drive the joint to the desired position while respecting kinematic
constraints. More tasks such as end-effector tracking can be added to the QP from within
the application without relaunching the simulation.

\section{Grasping Demonstration}

MuJoCo has been used widely for simulating object grasping and manipulation using
conventional model-based controllers and controllers based on learned policies
\cite{andrychowicz2020learning, kiatos2020geometric, kumar2015mujoco}.
The soft constraint model used in the library increases the stability of the
simulation while maintaining the realism.

To demonstrate the practicality of our proposed framework, we develop a simple FSM
controller using the HRP-5P humanoid robot for a simple demonstration of grasping an object placed
on the top of a table. For this demonstration, we choose HRP-5P as the robot platform due to the
wide size of its grippers required to grasp the box object, and due to prior knowledge
of tuned hyperparameters required for locomotion using the LIPM-Walking controller.
The FSM controller includes states for walking from the start position to
the edge of the table using the LIPM Walking controller \cite{caron2019stair}, hand retargetting,
and opening and closing of grippers to grasp and then lift up the desired object. \autoref{figure:fsm}
shows several keyframes of the implemented motion.
The robot starts in the ``Initial" state where the only task is to hold the current posture, equal
to the predefined ``half-sitting" stance. The controller then automatically enters the ``DemoFSM"
where the stabilizer tasks is executed in parallel to all other tasks. ``DemoFSM" begins
with walking over a predefined footstep plan to reach the table and perform ``Standing".
Then, the ``GraspStance" state adds a 6D end-effector transform task to place the right hand
over the table and near the object.  The remainder of the states for opening the gripper, reaching
the object, closing and then lifting up the object, are self-explanatory.
All state transitions occur automatically according to the completion criteria defined for each state.

Note that, in this particular example, \textit{mc-rtc} is not aware of the environment object and hence,
a collision between the robot and the table is possible. If the \textit{table} object is added to
\textit{mc-rtc}, the QP solver can be used to compute a collision-free trajectory.

The full demonstration can be seen in the attached video submission. The controller code with
installation and usage instruction is made available on GitHub.




\section{CONCLUSION}
Through this work, we developed and publicly released \textit{mc-mujoco} --- a framework
for simulating essentially \textit{any} arbitrary robot in the MuJoCo physics simulator
with FSM controllers expressed in \textit{mc-rtc}. \textit{mc-rtc} is an already existing
popular control framework for developing, tuning, evaluating controllers for a wide variety
of robots and has been extensively used to control many real platforms.

Similar to other \textit{mc-rtc} interfaces (like \textit{mc-openrtm})
\textit{mc-mujoco} is also developed in a transparent manner, meaning
that the same controller code that is used for verification in the simulation, can also
be used for real robot deployment using other communication interfaces. This is beneficial
for researchers and developers because it reduces the chances for inadvertently
introducing bugs, and maintains the validity or authenticity of the simulated behavior.

Elimination of binding dependency to ROS or other third party applications for graphically
acquiring user input is another strong feature of our proposed software.


We hope that the release of \textit{mc-mujoco} will facilitate research within the
robotics community for simulating robots and evaluating various controllers. We plan
to actively maintain the GitHub repository and introduce new and exciting features in
the future.

\section*{Acknowledgements}
The authors thank all members of JRL for the insightful discussions that led
to the production of this work. We also thank the users of \textit{mc-mujoco}
on GitHub. This work was partially supported by JST SPRING, Grant Number JPMJSP2124.

\balance
\bibliographystyle{IEEEtran}
\bibliography{IEEEabrv,bibliography.bib}

\begin{thebibliography}{10}
\providecommand{\url}[1]{#1}
\csname url@rmstyle\endcsname
\providecommand{\newblock}{\relax}
\providecommand{\bibinfo}[2]{#2}
\providecommand\BIBentrySTDinterwordspacing{\spaceskip=0pt\relax}
\providecommand\BIBentryALTinterwordstretchfactor{4}
\providecommand\BIBentryALTinterwordspacing{\spaceskip=\fontdimen2\font plus
\BIBentryALTinterwordstretchfactor\fontdimen3\font minus
  \fontdimen4\font\relax}
\providecommand\BIBforeignlanguage[2]{{%
\expandafter\ifx\csname l@#1\endcsname\relax
\typeout{** WARNING: IEEEtran.bst: No hyphenation pattern has been}%
\typeout{** loaded for the language `#1'. Using the pattern for}%
\typeout{** the default language instead.}%
\else
\language=\csname l@#1\endcsname
\fi
#2}}

\bibitem{todorov2012mujoco}
E.~Todorov, T.~Erez, and Y.~Tassa, ``Mujoco: A physics engine for model-based
  control,'' in \emph{2012 IEEE/RSJ international conference on intelligent
  robots and systems}.\hskip 1em plus 0.5em minus 0.4em\relax IEEE, 2012, pp.
  5026--5033.

\bibitem{mujocoblog}
\BIBentryALTinterwordspacing
Y.~Tassa and S.~Tunyasuvunakool, 2022. [Online]. Available:
  \url{https://www.deepmind.com/blog/open-sourcing-mujoco}
\BIBentrySTDinterwordspacing

\bibitem{erez2015simulation}
T.~Erez, Y.~Tassa, and E.~Todorov, ``Simulation tools for model-based robotics:
  Comparison of bullet, havok, mujoco, ode and physx,'' in \emph{2015 IEEE
  international conference on robotics and automation (ICRA)}.\hskip 1em plus
  0.5em minus 0.4em\relax IEEE, 2015, pp. 4397--4404.

\bibitem{escande2014hierarchical}
A.~Escande, N.~Mansard, and P.-B. Wieber, ``Hierarchical quadratic programming:
  Fast online humanoid-robot motion generation,'' \emph{The International
  Journal of Robotics Research}, vol.~33, no.~7, pp. 1006--1028, 2014.

\bibitem{cisneros2019qp}
R.~Cisneros, M.~Benallegue, M.~Morisawa, and F.~Kanehiro, ``Qp-based task-space
  hybrid/parallel control for multi-contact motion in a torque-controlled
  humanoid robot,'' in \emph{2019 IEEE-RAS 19th International Conference on
  Humanoid Robots (Humanoids)}.\hskip 1em plus 0.5em minus 0.4em\relax IEEE,
  2019, pp. 663--670.

\bibitem{ando2005rt}
N.~Ando, T.~Suehiro, K.~Kitagaki, T.~Kotoku, and W.-K. Yoon, ``Rt-component
  object model in rt-middleware-distributed component middleware for rt (robot
  technology),'' in \emph{2005 International Symposium on Computational
  Intelligence in Robotics and Automation}.\hskip 1em plus 0.5em minus
  0.4em\relax IEEE, 2005, pp. 457--462.

\bibitem{nakaoka2012choreonoid}
S.~Nakaoka, ``Choreonoid: Extensible virtual robot environment built on an
  integrated gui framework,'' in \emph{2012 IEEE/SICE International Symposium
  on System Integration (SII)}.\hskip 1em plus 0.5em minus 0.4em\relax IEEE,
  2012, pp. 79--85.

\bibitem{bolotnikova2021task}
A.~Bolotnikova, P.~Gergondet, A.~Tanguy, S.~Courtois, and A.~Kheddar,
  ``Task-space control interface for softbank humanoid robots and its
  human-robot interaction applications,'' in \emph{2021 IEEE/SICE International
  Symposium on System Integration (SII)}.\hskip 1em plus 0.5em minus
  0.4em\relax IEEE, 2021, pp. 560--565.

\bibitem{imgui}
\BIBentryALTinterwordspacing
O.~Cornut. [Online]. Available: \url{https://github.com/ocornut/imgui/}
\BIBentrySTDinterwordspacing

\bibitem{andrychowicz2020learning}
O.~M. Andrychowicz, B.~Baker, M.~Chociej, R.~Jozefowicz, B.~McGrew,
  J.~Pachocki, A.~Petron, M.~Plappert, G.~Powell, A.~Ray, \emph{et~al.},
  ``Learning dexterous in-hand manipulation,'' \emph{The International Journal
  of Robotics Research}, vol.~39, no.~1, pp. 3--20, 2020.

\bibitem{kiatos2020geometric}
M.~Kiatos, S.~Malassiotis, and I.~Sarantopoulos, ``A geometric approach for
  grasping unknown objects with multifingered hands,'' \emph{IEEE Transactions
  on Robotics}, vol.~37, no.~3, pp. 735--746, 2020.

\bibitem{kumar2015mujoco}
V.~Kumar and E.~Todorov, ``Mujoco haptix: A virtual reality system for hand
  manipulation,'' in \emph{2015 IEEE-RAS 15th International Conference on
  Humanoid Robots (Humanoids)}.\hskip 1em plus 0.5em minus 0.4em\relax IEEE,
  2015, pp. 657--663.

\bibitem{caron2019stair}
S.~Caron, A.~Kheddar, and O.~Tempier, ``Stair climbing stabilization of the
  hrp-4 humanoid robot using whole-body admittance control,'' in \emph{2019
  International Conference on Robotics and Automation (ICRA)}.\hskip 1em plus
  0.5em minus 0.4em\relax IEEE, 2019, pp. 277--283.

\end{thebibliography}
\end{document}